\documentclass[sigconf]{acmart}
\renewcommand\footnotetextcopyrightpermission[1]{} %remove copyright
\AtBeginDocument{%
  }

\setcopyright{acmlicensed}
\copyrightyear{2026}
\acmYear{2026}
\acmDOI{XXXXXXX.XXXXXXX}
\acmConference[MM]{The xxth ACM International Conference on Multimedia}{xx--xx xx xxxx}{xxx}
\acmISBN{xxxx}
\acmSubmissionID{7907}

\usepackage{gradient-text}
\usepackage{xspace}
\usepackage{booktabs}
\usepackage{multirow}
\usepackage{graphicx}
\usepackage[table]{xcolor}
\usepackage[most]{tcolorbox}
\newcommand{\ours}{\gradientRGB{CineCap}{13,118,128}{242, 133, 34}\xspace}
\newtcolorbox{findingbox}[1][Findings]{
    enhanced,
    sharp corners, % 取消圆角，纯直角更显硬朗和专业
    colback=black!2, % 背景设为 2% 的极浅灰，几乎为白，比纯白稍有层次
    colframe=black!75, % 边框设为深灰色（75%黑），比纯黑色柔和，不刺眼
    boxrule=0.8pt, % 边框粗细适中
    coltitle=black!85, % 标题文字颜色为更深的灰色
    fonttitle=\bfseries\itshape, % 标题使用加粗斜体，学术感更强
    attach boxed title to top left={xshift=1em, yshift=-\tcboxedtitleheight/2}, % 将标题卡在上方边框的左侧
    boxed title style={
        sharp corners,
        colframe=white, % 标题框的边框为白色
        colback=white,  % 标题框的背景为白色（用于遮挡主边框的线条）
        boxrule=0pt,    % 隐藏标题框自身的边框线
        left=2pt, right=2pt, top=0pt, bottom=0pt % 标题文字前后的留白
    },
    title=#1, % 支持自定义标题，默认是 Findings
    left=6pt, right=6pt, top=10pt, bottom=6pt % 框内正文留白
}
\begin{document}

\title{\ours: Structured Reasoning with Spatio-Temporal Anchors for Cinematographic Video Captioning}

\author{Xinyu Mao}
\authornote{Both authors contributed equally to this research.}
\authornote{This work was conducted during the author’s internship at Kling Team, Kuaishou Technology.}
% \author{Yuhui Zeng}
\affiliation{%
  \institution{The Chinese University of Hong Kong}
  \country{HKSAR}
}

\author{Yuhui Zeng}
\authornotemark[1]
\authornotemark[2]
% \author{Yuhui Zeng}
\affiliation{%
  \institution{Xiamen University}
  \country{China}
}

\author{Xiaokun Liu}
\authornote{Project Leader.}
% \author{Yuhui Zeng}
\affiliation{%
  \institution{Kling Team, Kuaishou Technology}
  \country{China}
}

\author{Wenyu Qin}
% \author{Yuhui Zeng}
\affiliation{%
  \institution{Kling Team, Kuaishou Technology}
  \country{China}
}

\author{Meng Wang}
% \author{Yuhui Zeng}
\affiliation{%
  \institution{Kling Team, Kuaishou Technology}
  \country{China}
}

\author{Xin Tao}
% \author{Yuhui Zeng}
\affiliation{%
  \institution{Kling Team, Kuaishou Technology}
  \country{China}
}

\author{Pengfei Wan}
% \author{Yuhui Zeng}
\affiliation{%
  \institution{Kling Team, Kuaishou Technology}
  \country{China}
}

\author{Xiaohan Xing}
% \author{Yuhui Zeng}
\affiliation{%
  \institution{National University of Singapore}
  \country{Singapore}
}

\author{Max Meng}
\authornote{Corresponding author. email: max.meng@ieee.org}
% \author{Yuhui Zeng}
\affiliation{%
  \institution{Southern University of Science and Technology}
  \country{China}
}
% \affiliation{%

% \affiliation{%

% \affiliation{%

% \affiliation{%

% \affiliation{%

% \affiliation{%

% \affiliation{%

\renewcommand{\shortauthors}{Mao et al.}

\begin{abstract}
Cinematographic captioning aims to describe how a video is filmed using professional film-language concepts such as camera movement, shot size, depth of field, composition, and shooting angle. This capability is important for fine-grained video understanding and controllable movie-quality video generation, yet remains underexplored in existing multimodal large language models. Unlike question-answering-based evaluation of cinematic understanding, cinematographic captioning requires a unified open-form description over multiple cinematographic dimensions. This task is challenging for two main reasons: the model must infer professional cinematographic concepts from subtle visual evidence, and it must generate captions that are both comprehensive and accurate. Accordingly, we propose \textbf{\ours}, a framework that combines structured reasoning with spatio-temporal anchors and reinforcement learning with comprehensiveness, accuracy, and gated coverage rewards. The former grounds professional cinematographic descriptions in explicit visual evidence and organizes them into compact atomic reasoning for supervised fine-tuning, while the latter improves the balance between descriptive completeness and factual correctness. In addition, we construct \textbf{CineCap Bench}, a benchmark of 472 manually annotated video-caption pairs for systematic evaluation. Extensive experiments show that \ours consistently outperforms strong proprietary and open-source baselines, establishing a new state of the art for cinematographic captioning. The code, model checkpoint, and benchmark are publicly available in our
\url{https://github.com/Hectormxy/CineCap.git}.
\end{abstract}


\begin{teaserfigure}
  \includegraphics[width=\textwidth]{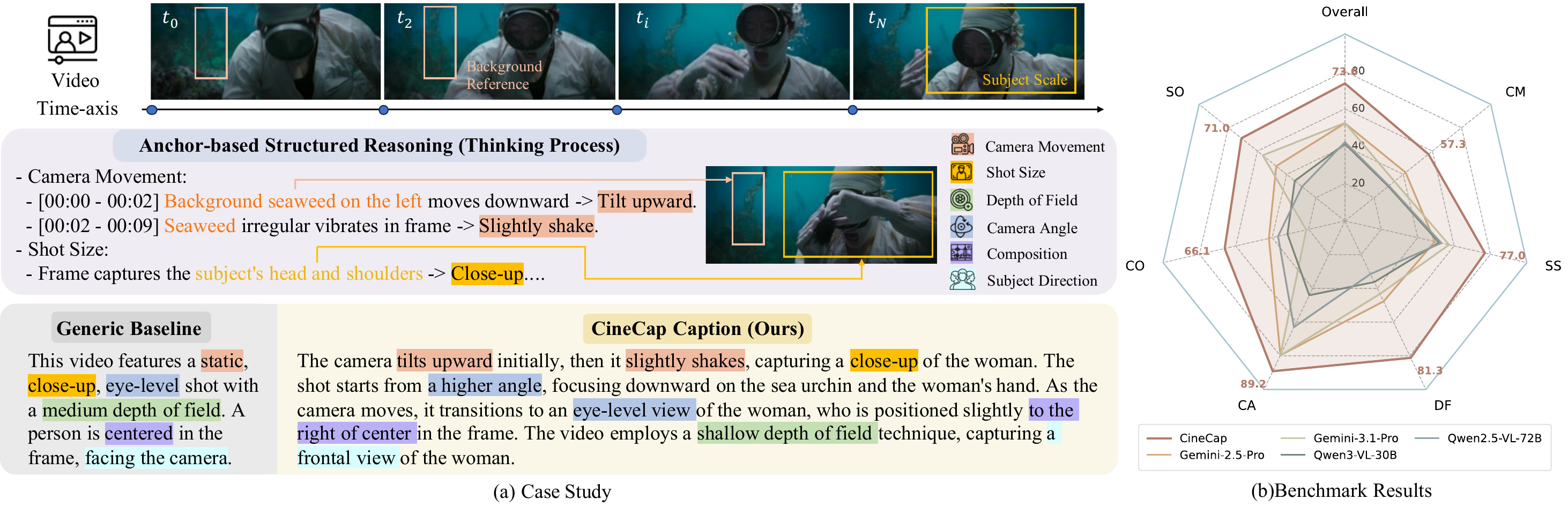}
  % \caption{Overview of our \ours. Given a video, the model first performs concise structured reasoning with spatio-temporal anchors to identify atomic cinematographic aspects, including camera movement, shot size, depth of field, camera angle, composition, and subject orientation. It then merges these aspect-level observations into a concise and fluent cinematographic dense caption.}
  \caption{Case study of our \ours and benchmark comparison. (a) Demonstration of the thinking process on a video sequence, where specific cinematic attributes like camera movement and shot size are inferred from spatial anchors. 
  \ours clearly provides more comprehensive and accurate camera movement descriptions than the generic baseline.
  (b) Radar chart showing benchmark results, indicating \ours outperforms existing vision-language models across multiple cinematic dimensions.}
  \label{fig:teaser}
\end{teaserfigure}


\maketitle

\section{Introduction}
% The cinematography of video, or camera-related cinematics \cite{chatterjee2025stable}, refers to the visual language that governs how visual content is filmed, organized, and presented over space and time \cite{bose2023movieclip, huang2020movienet, song2024moviechat, tapaswi2016movieqa, vicol2018moviegraphs}. Rather than merely recording scene content, cinematography shapes how visual information is structured, how attention is guided, and how motion, space, and narrative emphasis are conveyed to the viewer. As multimodal large language models (MLLMs) \cite{bai2025qwen2, Qwen2025Qwen3VL, Zhu2025InternVL3, OpenBMB2025MiniCPMV4_5} are increasingly expected to understand videos beyond coarse semantics, reliable cinematographic understanding is becoming an important capability. On the one hand, it requires the model not only to perceive scene content, but also to infer the state of the observer from changes in viewpoint, scale, and spatial relation, which is essential for three-dimensional spatial understanding. On the other hand, it provides an important foundation for controllable movie-quality video generation \cite{BarTal2024Lumiere,Zhang2023ControlVideo,Ma2024Latte,Huang2025StepVideoTI2V}, where accurate modeling of camera-related attributes is required to produce professional visual outputs.

The cinematography of video, also referred to as camera-related cinematics \cite{chatterjee2025stable}, denotes the visual language governing the manner in which visual content is filmed, organized, and presented across spatial and temporal dimensions \cite{bose2023movieclip, huang2020movienet, song2024moviechat, tapaswi2016movieqa, vicol2018moviegraphs}. Beyond the mere recording of scene content, cinematography shapes the structuring of visual information, directs viewer attention, and conveys motion, spatial relationships, and narrative emphasis. As multimodal large language models (MLLMs) \cite{bai2025qwen2, Qwen2025Qwen3VL, Zhu2025InternVL3, OpenBMB2025MiniCPMV4_5} are increasingly expected to understand videos at a level surpassing coarse semantic interpretation, reliable cinematographic understanding emerges as a crucial capability. This not only necessitates perception of scene content but also requires inference of the observer’s state via changes in viewpoint, scale, and spatial relations, which is indispensable for three-dimensional spatial comprehension. Moreover, it provides a critical foundation for controllable generation of movie-quality videos \cite{BarTal2024Lumiere,Zhang2023ControlVideo,Ma2024Latte,Huang2025StepVideoTI2V}, where precise modeling of camera-related attributes is essential to produce professional visual outputs.

Recently, cinematic understanding has garnered increasing attention within MLLM research \cite{liu2025shotbench, wang2025cinetechbench, lin2025towards, wu2026camreasoner}. One research direction formulates the problem as visual question answering, encompassing classification and multiple-choice formats \cite{liu2025shotbench, tang2025vidcomposition}, where models predict predefined cinematic concepts from limited candidate options. While convenient for benchmarking, these approaches primarily evaluate constrained recognition capabilities and do not require the model to articulate a unified description of how a video is filmed. Another direction addresses caption generation \cite{yao2026timechat} but typically restricts itself to limited factors, most commonly camera motion. For example, CamReasoner \cite{wu2026camreasoner} emphasizes camera motion understanding in a VQA-style setting rather than producing joint descriptions of broader cinematographic attributes. In contrast, our work focuses on cinematographic captioning, aiming to generate open-form descriptions that jointly encompass six key dimensions: camera movement, shot size, shooting angle, depth of field, composition, and subject orientation. This task is essential because cinematographic understanding in practice is inherently multi-dimensional \cite{chatterjee2025stable}, and isolated prediction of individual factors fails to fully capture how visual presentation is constructed. Compared to constrained recognition tasks, cinematographic captioning offers a more comprehensive generative formulation of cinematic understanding and a practical setting for end-to-end video-cinematography alignment.

% Despite its importance, cinematographic captioning remains a challenging task. First, it requires fine-grained understanding of professional cinematographic concepts that are often visually subtle and easily confused. For example, the model must distinguish camera motion from subject motion, and differentiate between closely related categories such as close-up and medium close-up. Second, cinematographic patterns within a video are often temporally compositional rather than static. A single video may contain compound camera behaviors, where the first segment exhibits one type of camera movement and the later segment shifts to another, as illustrated in Fig. \ref{fig:teaser}. This requires the model to capture temporally evolving cinematographic structure, rather than assigning a single label to the entire video. Third, as a dense captioning task \cite{chen2025avocado, yuan2025tarsier2, meng2025videocap}, cinematographic captioning requires the model not only to be accurate, but also to be comprehensive. It must faithfully cover multiple cinematographic dimensions and further integrate them into a coherent and fluent description.
Despite its significance, cinematographic captioning remains a challenging task. First, it requires fine-grained understanding of professional cinematographic concepts that are often visually subtle and prone to confusion. For instance, the model must distinguish between camera motion and subject motion, as well as differentiate closely related categories such as close-up and medium close-up. Second, cinematographic patterns within a video typically exhibit temporal compositionality rather than static characteristics. A single video may present compound camera behaviors, where one segment displays a particular type of camera movement and a subsequent segment shifts to another, as illustrated in Fig. \ref{fig:teaser}. This necessitates modeling of temporally evolving cinematographic structures instead of assigning a single label to the entire video. Third, as a dense captioning task \cite{chen2025avocado, yuan2025tarsier2, meng2025videocap}, cinematographic captioning requires the model to be not only accurate but also comprehensive. The description must faithfully cover multiple cinematographic dimensions and integrate them into a coherent and fluent narrative.

% To address these challenges, we propose \ours, a framework that grounds caption generation in explicit visual evidence through Structured Reasoning with Spatio-Temporal Anchors. Specifically, we introduce spatial anchors to infer professional cinematographic concepts from observable visual cues, and temporal anchors to associate dynamic aspects with specific timestamps. For example, in Fig. \ref{fig:teaser}, the model identifies a tilt-up during 00:00--00:02 from the downward motion of background seaweed, and a shaking phase during 00:02--00:09 from its irregular entry into and exit from the frame. To support unified description over multiple cinematographic dimensions, we build atomic structured chain-of-thought (CoT) data for supervised fine-tuning. We then introduce reinforcement learning with atomic caption evaluation using Group Relative Policy Optimization \cite{Guo2025DeepSeekR1IncentivizingReasoning} (GRPO), where an LLM-as-a-Judge provides a comprehensiveness score $s_{\mathrm{cmp}}$ and an accuracy score $s_{\mathrm{acc}}$. However, optimizing only $s_{\mathrm{cmp}}$ and $s_{\mathrm{acc}}$ in GRPO leads to insufficient coverage in practice. To address the trade-off between comprehensiveness and accuracy, where joint optimization of $s_{\mathrm{cmp}}$ and $s_{\mathrm{acc}}$ improves accuracy but degrades comprehensiveness, we further propose an atomic coverage reward, which constrains the number of described atomic aspects rather than caption length, thereby better balancing the two objectives.
To address these challenges, we propose \ours, a framework that grounds caption generation in explicit visual evidence via Structured Reasoning with Spatio-Temporal Anchors. Specifically, spatial anchors are introduced to infer professional cinematographic concepts from observable visual cues, while temporal anchors associate dynamic aspects with specific timestamps. For example, in Fig. \ref{fig:teaser}, the model detects a tilt-up during 00:00--00:02 from the downward motion of background seaweed and identifies a shaking phase between 00:02--00:09 by observing its irregular entry and exit from the frame. To support unified descriptions across multiple cinematographic dimensions, we construct atomic structured chain-of-thought (CoT) data for supervised fine-tuning. Subsequently, reinforcement learning is employed with atomic caption evaluation using Group Relative Policy Optimization \cite{Guo2025DeepSeekR1IncentivizingReasoning} (GRPO), where an LLM-as-a-Judge provides a comprehensiveness score $s_{\mathrm{cmp}}$ and an accuracy score $s_{\mathrm{acc}}$. However, sole optimization of $s_{\mathrm{cmp}}$ and $s_{\mathrm{acc}}$ within GRPO leads to insufficient coverage in practice. To address this trade-off, where joint optimization improves accuracy at the expense of comprehensiveness, we propose an atomic coverage reward that constrains the number of described atomic aspects rather than caption length, thus better balancing the two objectives.

% To systematically evaluate cinematic captioning quality, we construct CineCap Bench, a benchmark of 472 manually annotated video-caption pairs collected from public film datasets and YouTube videos. We evaluate captions at both the aspect level and the overall level, measuring comprehensiveness and accuracy across multiple cinematographic dimensions. Experimental results show that CineCap outperforms both closed-source models and a wide range of open-source baselines, establishing new state-of-the-art performance. These results confirm the effectiveness of our spatio-temporal anchor-based structured reasoning and reinforcement learning framework.
To systematically evaluate cinematic captioning quality, we build CineCap Bench, a benchmark comprising 472 manually annotated video-caption pairs sourced from public film datasets and YouTube videos. We assess captions both at the aspect level and overall, measuring comprehensiveness and accuracy across multiple cinematographic dimensions. Experimental results demonstrate that CineCap surpasses both closed-source models and an array of open-source baselines, establishing new state-of-the-art performance. These findings validate the efficacy of our spatio-temporal anchor-based structured reasoning and reinforcement learning framework.

% Our key contributions can be summarized as follows:
% \begin{itemize}
%     \item We propose \ours, the first unified framework for multi-dimensional cinematographic captioning, which generates joint descriptions of diverse cinematographic attributes in videos.
%     \item We introduce \textbf{spatio-temporal anchor-based structured reasoning} and \textbf{atomic coverage reward}, which improve compound professional concept understanding and the balance between description comprehensiveness and accuracy.
%     \item We construct \textbf{CineCap Bench} and conduct systematic comparisons with both open-source and closed-source models. Our method achieves state-of-the-art performance, validating the effectiveness of our framework.
% \end{itemize}

Our key contributions can be summarized as follows:
\begin{itemize}
    \item We introduce a novel perspective for addressing multi-dimensional cinematographic understanding by grounding caption generation in explicit spatio-temporal visual evidence, thereby tackling the inherent complexity of cinematic attribute interaction and temporal composition.
    \item We propose \ours, which introduces spatio-temporal anchor-based structured reasoning coupled with an atomic coverage reward mechanism to enhance fine-grained comprehension of professional cinematographic concepts and achieve a balanced trade-off between description comprehensiveness and accuracy.
    \item We construct \textbf{CineCap Bench}, the first comprehensive benchmark dataset for cinematic captioning, featuring 472 carefully annotated video-caption pairs covering diverse cinematographic aspects.
    \item We conduct extensive experiments showing that \ours consistently outperforms both open-source and proprietary baselines, achieving up to 32.41\% improvement in F1 evaluation.
\end{itemize}
\section{Related Work}
\subsection{Camera Related Video Analysis.} To enable cinematic video generation, understanding cinematography in videos \cite{lin2025towards,tang2025vidcomposition,liu2025shotbench,wang2025cinetechbench} has drawn growing attention. VidComposition \cite{tang2025vidcomposition} introduces a benchmark for evaluating the composition understanding ability of multimodal large language models (MLLMs) and comprehensively assesses 33 models on this task. Focusing specifically on camera motion, CameraBench \cite{lin2025towards} defines a rigorous taxonomy of camera motion primitives and collects a large set of expert-annotated video clips, demonstrating via SFT that MLLMs can acquire limited understanding of motion types and directions. CineTechBench \cite{wang2025cinetechbench} and ShotBench \cite{liu2025shotbench} further extend the evaluation scope to include composition, shot size, and depth of field, providing a more holistic benchmark for assessing cinematographic understanding in video models. However, most existing studies rely on multiple-choice evaluation \cite{rao2020unified,hu2025video,savardi2023cinescale2}.

\subsection{Reinforcement Learning for Vision Language Model.} Inspired by the success of reinforcement learning in large language models, recent studies have explored its application to multimodal large models \cite{zhou2024aligning, wang2024mdpo, zhang2025direct}. Vision-R1 \cite{huang2025vision} introduces a Progressive Thinking Suppression Training strategy combined with GRPO, effectively enhancing complex reasoning ability after cold-start training. VLM-R1 \cite{shen2025vlm} rigorously demonstrates the effectiveness and generalization of reinforcement learning on visual understanding tasks. R1-VL \cite{zhang2025r1} proposes Step-wise GRPO, enabling multimodal models to self-improve reasoning through simple yet dense step-wise rewards. For video understanding, Video-R1 \cite{feng2025video} creatively proposes T-GRPO, incorporating temporal modeling to promote explicit temporal reasoning; Video-RFT \cite{wang2025videorft} introduces a semantic-consistency reward to strengthen alignment between textual reasoning and visual evidence; and VideoChat-R1 \cite{feng2025video,yan2025videochatr15} systematically explores Reinforcement Fine-Tuning (RFT) with GRPO for video MLLMs. Unlike these works, our CineJudge targets video caption evaluation, which demands both temporal sensitivity and accurate assessment of video–caption alignment.

\subsection{Dense Captioning.} Dense captioning aims to generate multiple fine-grained descriptions for visual content and has been studied in both images and videos. DenseCap \cite{johnson2016densecap} first formulates dense captioning in images by jointly localizing salient regions and generating region-level descriptions. Dense-Captioning Events in Videos \cite{krishna2017dense} extends this setting to videos by detecting and describing multiple temporal events. Later work further improves dense video captioning through streamlined proposal-caption pipelines \cite{mun2019streamlined}. Reinforcement learning has also been widely explored to improve caption quality. SCST \cite{rennie2017self} optimizes non-differentiable caption metrics with policy gradients, while hierarchical reinforcement learning \cite{wang2018hrl} encourages more detailed video descriptions through multi-level decision making. More recently, CapRL \cite{xing2025caprl}, CCCaption \cite{tang2026cccaption}, and RubiCap \cite{huang2026rubicap} investigate fine-grained reward design for dense caption generation, focusing on utility, completeness and correctness, or structured rubric-based evaluation. However, these methods target general dense captioning rather than cinematic description. Our work instead studies dense captioning in the cinematographic domain, where the model must jointly describe multiple professional camera-related attributes.
\section{Task Formulation and Benchmark}

\begin{figure*}[h]
  \centering
  \includegraphics[width=\linewidth]{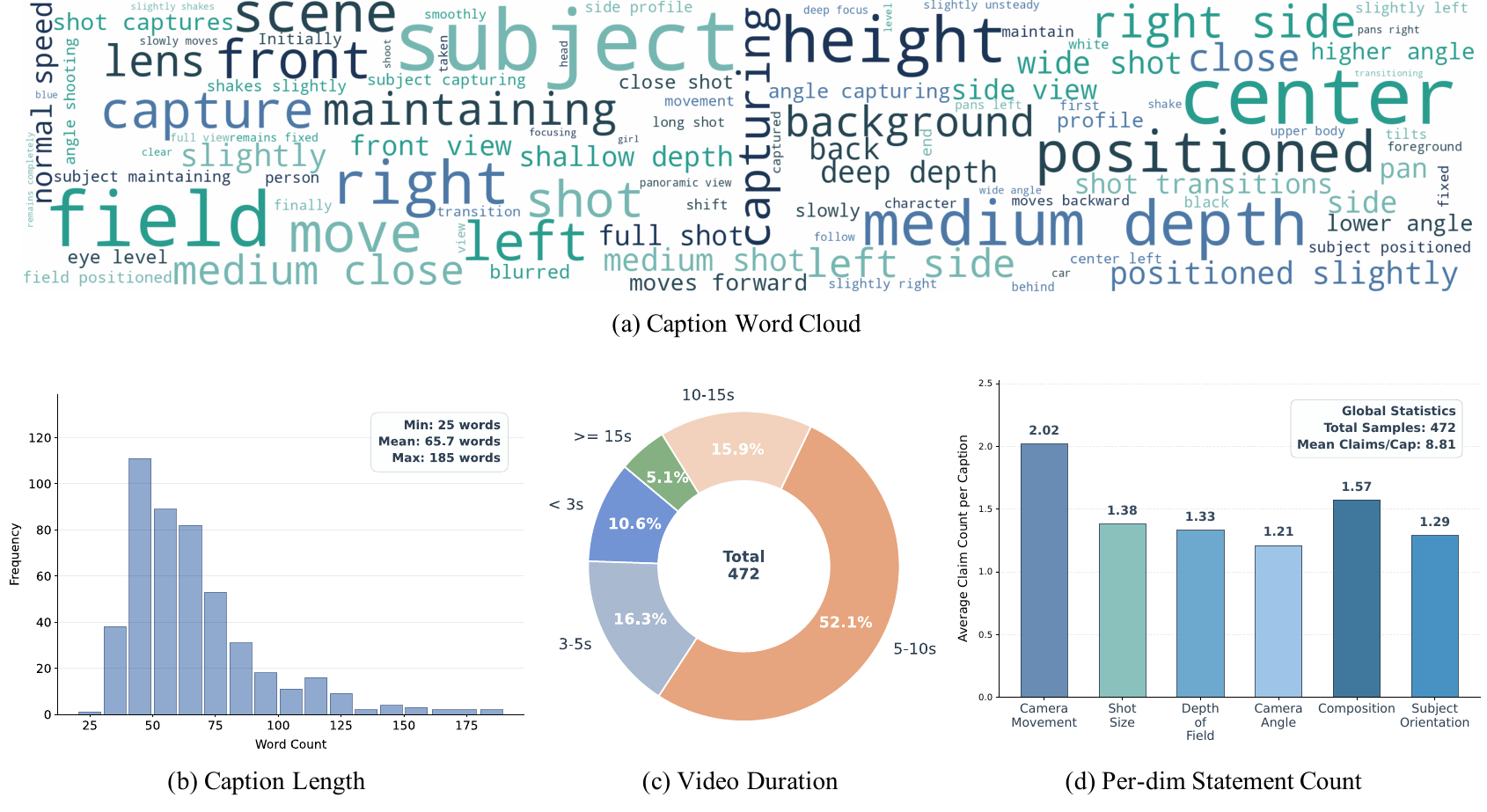}
  % \caption{Overview statistics of CineCap Bench. (a) Caption word cloud. (b) Caption length distribution, with lengths ranging from 25 to 185 words and an average of 65.7 words. (c) Video duration distribution, where most clips are between 5 and 10 seconds. (d) Average number of statements per caption for each cinematographic dimension.}
  \caption{Overview statistics of the CineCap Bench. (a) Word cloud of captions. (b) Distribution of caption lengths, ranging from 25 to 185 words with an average length of 65.7 words. (c) Distribution of video durations, where most clips are between 5 and 10 seconds. (d) Average number of statements per caption corresponding to each cinematographic dimension.}
  \label{fig:stat}
\end{figure*}

\subsection{Task Formulation}
% Given a video clip $v$, cinematographic captioning aims to generate a free-form caption $c$ that jointly describes its cinematographic properties. Specifically, the caption is expected to cover six dimensions: camera movement, shot size, shooting angle, depth of field, composition, and subject orientation. Different from standard video captioning, the task focuses not on what happens in the scene, but on how the scene is filmed and visually presented. Different from classification or multiple-choice settings, it requires unified generation over multiple dimensions rather than isolated prediction of predefined labels. The task therefore serves as a direct formulation of end-to-end video-cinematography alignment
Given a video clip v, the goal of cinematographic captioning is to generate a free-form caption c that comprehensively describes its cinematographic characteristics across six specific dimensions: camera movement, shot size, shooting angle, depth of field, composition, and subject orientation. Unlike standard video captioning, which focuses primarily on the events occurring within the scene, this task centers on how the scene is visually filmed and presented. Moreover, it diverges from typical classification or multiple-choice approaches by requiring a unified caption that covers multiple dimensions simultaneously, rather than producing isolated predictions for predefined labels. Thus, this task directly addresses the problem of aligning video content with cinematographic description in an end-to-end manner.

\subsection{Data Construction}
\noindent
\textbf{Data Source.}
% For cinematographic captioning, the quality of the source video is important, since fine-grained camera-related attributes require both sufficient visual clarity and rich cinematic expression to be reliably observed. We therefore collect our data from two sources: YouTube videos and public film content. To obtain clips with consistent cinematographic structure, we first apply \textit{PySceneDetect} \cite{pyscenedetect} to segment raw videos into single-shot clips. As our primary focus is camera-related cinematography, we begin by annotating each clip with a camera-motion category label. We then use these labels to balance the data distribution across motion types, and further annotate dense cinematographic captions on the balanced subset. This pipeline improves coverage of key camera-motion patterns and provides a more suitable data foundation for multi-dimensional cinematographic captioning.
For cinematographic captioning, the quality of source videos is critical since fine-grained camera-related attributes demand sufficient visual clarity and rich cinematic expression to be reliably perceived. To this end, data are collected from two sources: YouTube videos and publicly available film content. To obtain clips with consistent cinematographic structure, \textit{PySceneDetect} \cite{pyscenedetect} is first applied to segment raw videos into single-shot clips. Focusing primarily on camera-related cinematography, each clip is initially annotated with a camera-motion category label. These labels are then used to balance the data distribution across motion types, after which dense cinematographic captions are annotated on the balanced subset. This pipeline enhances coverage of key camera-motion patterns and establishes a more appropriate data foundation for multi-dimensional cinematographic captioning.

\par
\noindent
\textbf{Annotation Pipeline.}
% To ensure annotation quality and consistency, we recruit annotators with backgrounds in aesthetics or film-related fields, and require them to complete training and qualification tests before formal annotation. For each video clip, we adopt a semi-automatic pipeline: we first use the closed-source model Gemini 3 Pro \cite{gemini3pro} to generate an initial caption, and then ask annotators to revise it by correcting errors, adding missing details, removing unsupported content, and refining professional terminology. The revised caption is finally normalized into a unified output format. To guarantee data quality, we further apply a two-stage review process after annotation: a first-round review randomly checks about 30\% of the samples, followed by a second-round expert review on about 10\% of the samples. This multi-stage pipeline improves the accuracy, consistency, and professionalism of the resulting annotations.
To ensure annotation quality and consistency, annotators with backgrounds in aesthetics or film-related fields are recruited and required to complete training and qualification tests prior to formal annotation. For each video clip, a semi-automatic pipeline is adopted: an initial caption is generated using the closed-source model Gemini 3 Pro \cite{gemini3pro}, after which annotators revise the caption by correcting errors, adding missing details, removing unsupported content, and refining professional terminology. The revised caption is subsequently normalized into a unified output format. To guarantee data quality, a two-stage review process is implemented post-annotation: a first-round review randomly inspects approximately 30\% of the samples, followed by a second-round expert review on about 10\% of the samples. This multi-stage procedure improves the accuracy, consistency, and professionalism of the resulting annotations.

\par
\noindent
\textbf{Benchmark Statistics.}
% CineCap Bench contains 472 manually annotated video-caption pairs. As shown in Fig.~\ref{fig:stat}a, the caption word cloud is dominated by cinematography-oriented expressions related to framing, shot scale, viewpoint, and subject position, indicating that the benchmark captures professional visual presentation rather than generic scene semantics. As shown in Fig.~\ref{fig:stat}b-c, the benchmark further spans a wide range of video durations and caption lengths. Most clips are between 5 and 10 seconds, while the captions range from short descriptions to relatively long paragraphs, with an average length of 65.7 words. This variation indicates that cinematographic captioning requires flexible description granularity rather than a fixed-length output.
CineCap Bench comprises 472 manually annotated video-caption pairs. As shown in Fig.~\ref{fig:stat}a, the caption word cloud predominantly features cinematography-oriented expressions related to framing, shot scale, viewpoint, and subject position, indicating that the benchmark captures professional visual presentation rather than generic scene semantics. Figures~\ref{fig:stat}b--c illustrate that the benchmark covers a wide range of video durations and caption lengths. Most clips last between 5 and 10 seconds, while captions range from brief descriptions to relatively long paragraphs, with an average length of 65.7 words. This variability indicates that cinematographic captioning requires flexible description granularity rather than a fixed-length output.

% We further examine the distribution of statement counts across cinematographic dimensions, shown in Fig.~\ref{fig:stat}d. Different dimensions exhibit different annotation patterns, showing that cinematographic description is inherently multi-dimensional and compositional. In particular, some dimensions, such as camera movement and composition, often involve multiple statements within a single caption, reflecting the temporally evolving and compound nature of cinematographic expression. These statistics demonstrate that CineCap Bench presents realistic variation in both temporal structure and descriptive granularity, making it suitable for evaluating whether a model can generate captions that are both comprehensive and accurate.
Further examination of the distribution of statement counts across cinematographic dimensions is presented in Fig.~\ref{fig:stat}d. Different dimensions exhibit distinct annotation patterns, demonstrating that cinematographic description is inherently multi-dimensional and compositional. In particular, certain dimensions, such as camera movement and composition, frequently involve multiple statements within a single caption, reflecting the temporally evolving and compound nature of cinematographic expression. These statistics confirm that CineCap Bench presents realistic variation in both temporal structure and descriptive granularity, making it suitable for evaluating whether a model can generate captions that are both comprehensive and accurate.

\subsection{Evaluation Protocol}
\label{eval}
% We evaluate cinematographic captioning from two perspectives: \textbf{comprehensiveness} and \textbf{accuracy}. Comprehensiveness measures whether a caption sufficiently covers the cinematographic attributes expressed in the video, while accuracy measures whether the described attributes are visually correct. Both criteria are necessary for this task. A caption may be accurate but incomplete if it only describes a subset of the relevant dimensions, and it may be comprehensive but unreliable if it introduces unsupported or incorrect descriptions.
Cinematographic captioning is evaluated from two perspectives: \textbf{comprehensiveness} and \textbf{accuracy}. Comprehensiveness assesses whether a caption sufficiently covers the cinematographic attributes expressed in the video, whereas accuracy evaluates whether the described attributes are visually correct. Both criteria are essential. A caption may be accurate but incomplete if it describes only a subset of the relevant dimensions, and it may be comprehensive yet unreliable if it includes unsupported or incorrect descriptions.

% To reflect the multi-dimensional nature of cinematographic captioning, we perform evaluation at both the \textbf{aspect level} and the \textbf{overall level}. The aspect-level evaluation measures whether the generated caption provides sufficiently complete and factually correct descriptions for each of the six cinematographic dimensions, including camera movement, shot size, shooting angle, depth of field, composition, and subject orientation, while the overall-level evaluation assesses whether the caption as a whole offers a globally comprehensive and accurate account of how the video is filmed. We report both comprehensiveness and accuracy at these two levels to evaluate not only fine-grained performance on individual factors but also the quality of the caption as a unified dense description.
To capture the multi-dimensional nature of the task, evaluation is performed at both the \textbf{aspect level} and the \textbf{overall level}. The aspect-level evaluation measures whether the generated caption provides sufficiently complete and factually accurate descriptions for each of the six cinematographic dimensions: camera movement, shot size, shooting angle, depth of field, composition, and subject orientation. The overall-level evaluation assesses whether the caption as a whole offers a globally comprehensive and accurate account of how the video is filmed. We report both comprehensiveness and accuracy at these two levels to assess not only fine-grained performance on individual factors but also the quality of the caption as a unified dense description.
\section{Method}
\begin{figure*}[h]
  \centering
  \includegraphics[width=\linewidth]{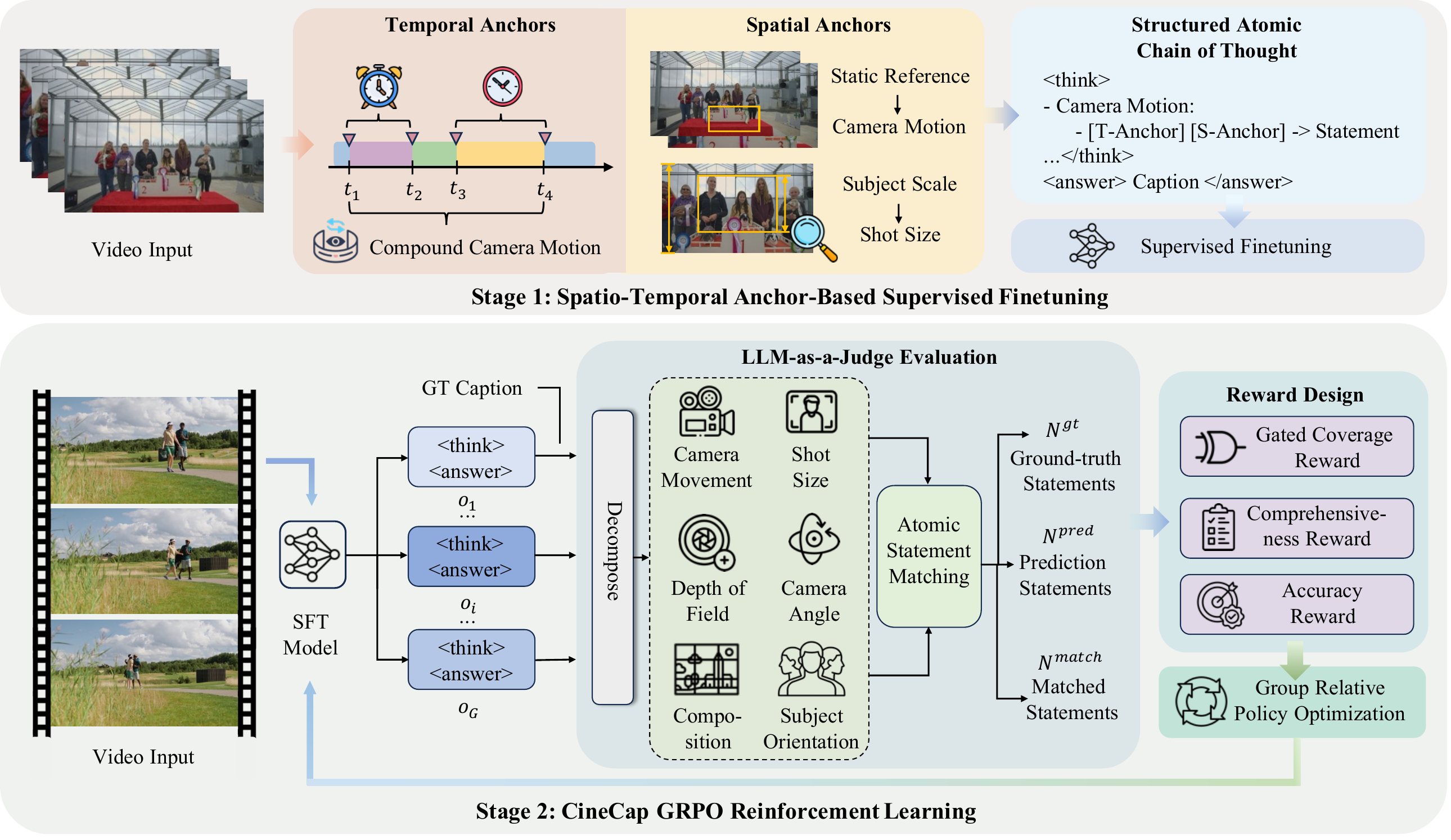}
  % \caption{Overview of \ours. Stage 1 uses spatio-temporal anchors to construct atomic CoT supervision. Stage 2 applies GRPO with comprehensiveness, accuracy, and gated coverage rewards.}
  \caption{Overview of \ours. Stage 1 employs spatio-temporal anchors to construct atomic CoT supervision. Stage 2 utilizes GRPO with rewards designed for comprehensiveness, accuracy, and gated coverage.}
  \label{fig:main}
\end{figure*}
\subsection{Overview}
% Given a video clip, our goal is to generate a dense cinematographic caption jointly describing multiple camera-related attributes in a unified paragraph. This task requires the model not only to infer professional cinematographic concepts from explicit visual evidence, but also to balance comprehensiveness and accuracy in open-form generation. To this end, we propose a two-stage framework. In the first stage, we introduce \textbf{Spatio-Temporal Anchor-Based Structured Reasoning}, which organizes visual evidence into structured reasoning for multi-dimensional cinematographic description, and use it to construct atomic CoT supervision for supervised fine-tuning. In the second stage, we further apply \textbf{GRPO} with comprehensiveness, accuracy, and coverage rewards to improve the quality of generated captions. The overall pipeline is shown in Fig.
Given a video clip, the objective is to generate a dense cinematographic caption that jointly describes multiple camera-related attributes within a unified paragraph. This task requires the model not only to infer professional cinematographic concepts from explicit visual evidence but also to balance comprehensiveness and accuracy in an open-form generation setting. To address these challenges, we propose a two-stage framework. In the first stage, \textbf{Spatio-Temporal Anchor-Based Structured Reasoning} is introduced to organize visual evidence into structured reasoning for multi-dimensional cinematographic description, which is then used to construct atomic chain-of-thought (CoT) supervision for supervised fine-tuning. In the second stage, \textbf{GRPO} is applied with rewards targeting comprehensiveness, accuracy, and coverage to enhance the quality of the generated captions. The overall pipeline is illustrated in Fig.~\ref{fig:main}.

% \subsection{Spatio-Temporal Anchor-Based Structured Reasoning}
\subsection{Spatio-Temporal Anchor-Based Reasoning}
% We observe that cinematographic attributes are not directly observable as discrete labels from video alone. Camera motion is usually perceived relative to reference objects in the scene, and professional cinematographic concepts require explicit interpretation of visual evidence. Furthermore, a video clip may contain compound cinematographic patterns, where camera movement changes over time and other dimensions, such as composition and subject orientation, evolve accordingly. These characteristics make single-shot global description insufficient for cinematographic captioning. Based on this observation, we propose Spatio-Temporal Anchor-Based Structured Reasoning, which introduces \textbf{spatial anchors} to ground professional concepts in scene evidence and \textbf{temporal anchors} to localize dynamically changing attributes.
Cinematographic attributes are not directly observable as discrete labels from video alone. Camera motion is typically perceived relative to reference objects within the scene, and professional cinematographic concepts require explicit interpretation of visual evidence. A video clip may contain compound cinematographic patterns, with camera movement evolving over time and other dimensions, such as composition and subject orientation, changing accordingly. These characteristics render single-shot global descriptions insufficient for cinematographic captioning. Based on this observation, we propose Spatio-Temporal Anchor-Based Structured Reasoning, which introduces \textbf{spatial anchors} to ground professional concepts in scene evidence, and \textbf{temporal anchors} to localize dynamically changing attributes.
\par
\noindent
\textbf{Spatial anchors.}
% Spatial anchors ground fine-grained cinematographic concepts in directly observable visual evidence. Instead of predicting abstract film-language terms holistically, the model first identifies concrete cues and then infers the corresponding attribute. For instance, camera motion is inferred from the positional change of static background references, which helps distinguish it from subject motion; subject scale indicates shot size; foreground-background sharpness informs depth of field. In this way, spatial anchors explicitly connect scene evidence with professional cinematographic terminology. All these cues are directly observable from the current frame or local visual content.
Spatial anchors ground fine-grained cinematographic concepts in directly observable visual evidence. Rather than predicting abstract film-language terms holistically, the model first identifies concrete cues and subsequently infers the corresponding attribute. For example, camera motion is inferred from positional changes of static background references, which distinguishes it from subject motion; subject scale indicates shot size; foreground-background sharpness informs depth of field. In this manner, spatial anchors explicitly link scene evidence with professional cinematographic terminology. All such cues are directly observable from the current frame or local visual content.

\par
\noindent
\textbf{Temporal anchors.}
% Temporal anchors are used to bind dynamic cinematographic attributes to localized temporal intervals. Formally, for a video clip, each dynamic attribute is represented as one or more anchors $\tau=[t_s,t_e]$, indicating the temporal region in which the attribute is expressed. This enables the model to describe compound cinematographic patterns within a single clip, rather than forcing a single global description. In practice, different intervals may correspond to different camera motions, while related dimensions such as composition and subject orientation can also change with the camera trajectory. Temporal anchors therefore provide the temporal structure needed for multi-stage and compound cinematographic reasoning. \\
By combining spatial and temporal anchors, our structured reasoning framework provides an intermediate representation between raw video signals and final language output. It enables the model to describe cinematographic attributes based on explicit evidence, while preserving the compositional structure needed for dense multi-dimensional captioning.
Temporal anchors bind dynamic cinematographic attributes to localized temporal intervals. Formally, for a video clip, each dynamic attribute is represented by one or more anchors $\tau = [t_s, t_e]$ indicating the temporal segment during which the attribute is exhibited. This enables the model to describe compound cinematographic patterns within a single clip rather than generating a single global description. In practice, different intervals may correspond to different camera motions, while related dimensions such as composition and subject orientation may also vary with the camera trajectory. Temporal anchors thus provide the temporal structure necessary for multi-stage and compound cinematographic reasoning.

By combining spatial and temporal anchors, the structured reasoning framework provides an intermediate representation bridging raw video signals and final language output. This facilitates description of cinematographic attributes based on explicit evidence while preserving the compositional structure required for dense multi-dimensional captioning.

\subsection{Supervised Fine-Tuning with Atomic CoT}

% We construct our supervised fine-tuning data in the form of \textbf{atomic chain-of-thought (CoT)} rather than unconstrained long-form reasoning. This design is motivated by two observations. First, recent video-language studies suggest that explicit CoT is not uniformly beneficial for all multimodal tasks; for some perception-heavy settings, direct answering can match or even outperform CoT despite its higher generation cost \cite{wang2026videoautor1}. Second, we observe that overly long reasoning chains in cinematographic captioning often introduce redundant statements and may accumulate errors across steps.
We construct supervised fine-tuning data in the form of \textbf{atomic chain-of-thought (CoT)} instead of unconstrained long-form reasoning. This design is motivated by two observations. First, recent video-language studies indicate that explicit CoT is not uniformly beneficial across all multimodal tasks; for some perception-intensive settings, direct answering can match or outperform CoT despite its higher generation cost \cite{wang2026videoautor1}. Second, overly long reasoning chains in cinematographic captioning often introduce redundant statements and accumulate errors across steps.

% Based on these observations, we do not impose explicit reasoning on every cinematographic aspect. Instead, we apply anchor-based reasoning only to attributes that genuinely require inference, especially camera motion and other temporally evolving patterns. In contrast, aspects such as subject orientation or composition are often directly observable from the current frame or local visual content, and are therefore supervised with direct statements rather than expanded reasoning chains.
In light of these observations, explicit reasoning is applied only to cinematographic attributes that require inference, notably camera motion and other temporally evolving patterns. Conversely, attributes such as subject orientation or composition are often directly observable from the current frame or local visual content, and are therefore supervised with direct statements rather than extended reasoning chains.

% Accordingly, we organize the supervision into an atomic structured format. For dynamic attributes that require temporal reasoning, each unit takes the form \texttt{[temporal anchor] [spatial anchor] $\rightarrow$ statement}, where \texttt{[spatial anchor]} is optional. For directly observable attributes, we use the statement alone. This design keeps the supervision compact while preserving explicit reasoning only where it is necessary, enabling the model to learn multi-dimensional cinematographic descriptions without relying on unnecessarily long CoT trajectories.
Accordingly, the supervision is organized into an atomic structured format. For dynamic attributes requiring temporal reasoning, each unit follows the form \texttt{[temporal anchor] [spatial anchor] $\rightarrow$ statement}, where \texttt{[spatial anchor]} is optional. For directly observable attributes, supervision consists of the statement alone. This design maintains concise supervision while preserving explicit reasoning where necessary, enabling the model to learn multi-dimensional cinematographic descriptions without relying on unnecessarily long CoT trajectories.

% To construct the supervision data, we start from manually annotated video-caption pairs and generate atomic CoT annotations conditioned on the ground-truth captions. The construction process recovers the visual evidence underlying each target description, yielding an evidence-to-answer CoT that better matches the evidence-grounded nature of cinematographic captioning. The generated reasoning traces are further reviewed by human annotators and a closed-source model to improve logical coherence and reduce reasoning or formatting errors. All samples are finally normalized into a unified \texttt{<think>...</think><answer>...</answer>} format, resulting in 80K video--CoT--caption training samples for supervised fine-tuning. More details of the construction pipeline and the exact prompts are provided in the supplementary material.
To construct the supervision data, we begin with manually annotated video-caption pairs and generate atomic CoT annotations conditioned on ground-truth captions. This construction recovers the visual evidence underpinning each target description, producing an evidence-to-answer CoT that aligns with the evidence-grounded nature of cinematographic captioning. The generated reasoning traces undergo review by human annotators and a closed-source model to enhance logical coherence and mitigate reasoning or formatting errors. Finally, all samples are normalized into a unified \texttt{<answer>...</answer>} format, yielding 80K video--CoT--caption training samples for supervised fine-tuning. Additional details regarding the construction pipeline and exact prompts are provided in the supplementary material.

\subsection{Fine-grained Reward Design}

% We further optimize the SFT model with Group Relative Policy Optimization (GRPO) to improve caption quality under the two core objectives of cinematographic captioning, namely comprehensiveness and accuracy. Given a sampled caption, we use an LLM-as-a-Judge to perform atomic evaluation following the same six-aspect decomposition as in our CoT supervision, including Camera Movement, Shot Size, Depth of Field, Camera Angle, Composition, and Subject Orientation. For each aspect \(d\), the judge decomposes both the ground-truth caption and the generated caption into atomic statements, and returns the corresponding numbers of ground-truth statements \(n_d^{\mathrm{gt}}\), predicted statements \(n_d^{\mathrm{pred}}\), and semantically matched statements \(n_d^{\mathrm{match}}\). We then aggregate them over all six aspects:

The supervised fine-tuned model is further optimized via Group Relative Policy Optimization (GRPO) to improve caption quality according to two fundamental objectives of cinematographic captioning: comprehensiveness and accuracy. Given a sampled caption, an LLM is employed as a judge to perform atomic evaluation following the same six-aspect decomposition employed in our CoT supervision, including Camera Movement, Shot Size, Depth of Field, Camera Angle, Composition, and Subject Orientation. For each aspect \(d\), the judge decomposes both the ground-truth and generated captions into atomic statements, producing counts of ground-truth statements \(n_d^{\mathrm{gt}}\) , predicted statements \(n_d^{\mathrm{pred}}\), and semantically matched statements \(n_d^{\mathrm{match}}\). These counts are aggregated over all six aspects:
\begin{equation}
N^{\mathrm{gt}}=\sum_{d=1}^{6} n_d^{\mathrm{gt}}, \qquad
N^{\mathrm{pred}}=\sum_{d=1}^{6} n_d^{\mathrm{pred}}, \qquad
N^{\mathrm{match}}=\sum_{d=1}^{6} n_d^{\mathrm{match}}.
\end{equation}
Based on these statistics, the comprehensiveness score and accuracy score are defined as
\begin{equation}
s_{\mathrm{cmp}} = \frac{N^{\mathrm{match}}}{\max(N^{\mathrm{gt}},1)},
\qquad
s_{\mathrm{acc}} = \frac{N^{\mathrm{match}}}{\max(N^{\mathrm{pred}},1)}.
\end{equation}
% Here, \(s_{\mathrm{cmp}}\) measures how much of the ground-truth cinematographic content is covered by the generated caption, while \(s_{\mathrm{acc}}\) measures how much of the generated content is factually correct.
Here, \(s_{\mathrm{cmp}}\) measures the proportion of ground-truth cinematographic content covered by the generated caption, while \(s_{\mathrm{acc}}\) measures the factual correctness of generated content.

% A natural choice is to combine \(s_{\mathrm{cmp}}\) and \(s_{\mathrm{acc}}\) directly as reinforcement signals. In practice, however, we observe that the gain in accuracy often dominates the overall reward. As a result, the model tends to improve \(s_{\mathrm{acc}}\) by producing shorter and more conservative captions, which in turn limits the improvement of comprehensiveness. To better balance these two objectives, we introduce an additional \textbf{coverage reward} that regularizes the overall number of described atomic statements. Unlike length-based regularization, our coverage reward is defined on atomic statement counts rather than caption length, based on the observation that a longer caption does not necessarily yield higher comprehensiveness; what matters is the coverage of key atomic cinematographic information.
A natural approach is to combine \(s_{\mathrm{cmp}}\) and \(s_{\mathrm{acc}}\) directly as reinforcement signals. However, in practice, improvement in accuracy tends to dominate the overall reward. Consequently, the model favors producing shorter and more conservative captions to increase \(s_{\mathrm{acc}}\), which limits gains in comprehensiveness. To balance these objectives, an additional \textbf{coverage reward} is introduced to regularize the overall count of described atomic statements. Unlike length-based regularization, this coverage reward is defined using atomic statement counts rather than caption length, motivated by the observation that a longer caption does not necessarily yield higher comprehensiveness; what matters is coverage of critical atomic cinematographic information.

Concretely, the coverage reward is defined as
\begin{equation}
r_{\mathrm{cov}}
=
-\min\left(1, \frac{|N^{\mathrm{gt}} - N^{\mathrm{pred}}|}{\max(N^{\mathrm{gt}}, 1)}\right).
\end{equation}
% This reward encourages the generated caption to match the target coverage at the atomic level. However, we further observe that directly introducing \(r_{\mathrm{cov}}\) can over-suppress the improvement of accuracy, since coverage control may encourage additional statements even when their correctness is not yet reliable. To address this issue, we apply a gated design and activate the coverage reward only when the caption has reached a sufficient accuracy level. Specifically, we define the gated coverage reward as
This reward encourages the generated caption to better match the target coverage at the atomic level. However, we observe that directly applying \(r_{\mathrm{cov}}\) can suppress accuracy improvement, since coverage control may encourage additional statements even when their correctness is not yet reliable. To mitigate this, a gated design is implemented, activating the coverage reward only when the caption attains sufficient accuracy. Specifically, the gated coverage reward is defined as
\begin{equation}
r_{\mathrm{cov}}^{\mathrm{gate}}
=
\mathbb{I}(s_{\mathrm{acc}}>\tau)\,
\left(
-\min\!\left(1,\frac{|N^{\mathrm{gt}}-N^{\mathrm{pred}}|}{\max(N^{\mathrm{gt}},1)}\right)
\right),
\end{equation}
where \(\mathbb{I}(\cdot)\) is the indicator function and \(\tau\) is an accuracy threshold.

For the \(i\)-th sampled response, we set
\begin{equation}
r_{\mathrm{cmp},i} = s_{\mathrm{cmp},i}, \qquad
r_{\mathrm{acc},i} = s_{\mathrm{acc},i},
\end{equation}
and define the final reward as
\begin{equation}
R_i = \lambda_{\mathrm{cmp}} r_{\mathrm{cmp},i} + \lambda_{\mathrm{acc}} r_{\mathrm{acc},i} + \lambda_{\mathrm{cov}} r_{\mathrm{cov},i}^{\mathrm{gate}}.
\end{equation}
\begin{table*}[!t]
\centering
\caption{Performance comparison of various models across different metrics. The best results are highlighted in \textbf{bold}. CM denotes Camera Movement. SS denotes Shot Size. DF denotes Depth of Field. CA denotes Camera Angle. CO denotes Composition. SO denotes Subject Orientation. Cmp denotes comprehensiveness. Acc denotes accuracy.}
\label{tab:main_result}
% 使用 resizebox 确保宽表格能自适应页面双栏的宽度
\resizebox{\textwidth}{!}{
\begin{tabular}{l ccccccccccccccc}
\toprule
\multirow{2}{*}{Model} & \multicolumn{2}{c}{CM} & \multicolumn{2}{c}{SS} & \multicolumn{2}{c}{DF} & \multicolumn{2}{c}{CA} & \multicolumn{2}{c}{CO} & \multicolumn{2}{c}{SO} & \multicolumn{3}{c}{Overall} \\
\cmidrule(lr){2-3} \cmidrule(lr){4-5} \cmidrule(lr){6-7} \cmidrule(lr){8-9} \cmidrule(lr){10-11} \cmidrule(lr){12-13} \cmidrule(lr){14-16}
 & Cmp & Acc & Cmp & Acc & Cmp & Acc & Cmp & Acc & Cmp & Acc & Cmp & Acc & Cmp & Acc & F1 \\
\midrule
\multicolumn{16}{l}{\textit{Proprietary models}} \\
\midrule
Gemini-2.5-Pro\cite{gemini25pro} & 40.23 & 43.09 & 51.07 & 47.92 & 48.11 & 47.37 & 80.92 & 78.61 & 40.24 & 43.55 & 48.90 & 45.44 & 52.03 & 52.53 & 52.28 \\
Gemini-3.1-Pro\cite{gemini31pro} & 29.22 & 40.50 & 54.34 & 60.52 & 43.01 & 43.59 & 79.00 & 80.60 & 19.98 & 22.76 & 54.78 & 58.01 & 46.31 & 60.48 & 52.45 \\
\midrule
\multicolumn{16}{l}{\textit{Open-source models ($>$10B)}} \\
\midrule
Qwen3-VL-30B\cite{Qwen2025Qwen3VL}   & 29.22 & 30.31 & 54.26 & 49.47 & 36.12 & 36.39 & 44.67 & 43.53 & 30.19 & 33.16 & 34.52 & 34.32 & 39.33 & 43.42 & 41.27 \\
Qwen2.5-VL-72B\cite{bai2025qwen2} & 28.42 & 32.23 & 53.19 & 52.83 & 31.92 & 31.11 & 65.65 & 60.58 & 35.71 & 37.93 & 28.70 & 30.18 & 41.02 & 43.39 & 42.17 \\
\midrule
\multicolumn{16}{l}{\textit{Open-source models ($<$10B)}} \\
\midrule
Tarsier-7B\cite{yuan2024tarsier}     & 8.06  & 13.11 & 37.57 & 35.51 & 27.38 & 28.28 & 15.81 & 14.92 & 17.83 & 21.47 & 22.95 & 25.23 & 22.64 & 31.05 & 26.19 \\
InternVL-3-8B\cite{Zhu2025InternVL3}  & 18.80 & 25.21 & 42.70 & 40.32 & 33.66 & 34.89 & 27.17 & 26.41 & 27.85 & 31.54 & 29.27 & 31.57 & 30.69 & 36.81 & 33.47 \\
LLaVa-OV-7B\cite{li2024llava}    & 8.94  & 13.51 & 28.21 & 25.98 & 19.50 & 20.04 & 11.66 & 10.86 & 16.14 & 17.98 & 19.79 & 21.10 & 19.30 & 31.40 & 23.91 \\
LLaVa-NV-7B\cite{zhang2024llavanextvideo}    & 8.11  & 12.89 & 56.68 & 34.91 & 23.23 & 23.71 & 30.37 & 25.44 & 24.31 & 29.91 & 20.23 & 22.02 & 28.05 & 30.14 & 29.06 \\
\rowcolor{gray!30}
Qwen3-VL-8B\cite{Qwen2025Qwen3VL}    & 34.48 & 31.66  & 45.49  & 38.52  & 36.76  & 33.40  & 61.12  & 51.14  & 33.31  & 36.73  & 44.19  & 41.21  & 44.09  & 38.60  & 41.16  \\
\midrule
\rowcolor{cyan!20}
\textbf{Ours} (CineCap) & \textbf{54.38} & \textbf{60.50} & \textbf{76.51} & \textbf{77.44} & \textbf{79.68} & \textbf{82.91} & \textbf{90.06} & \textbf{88.35} & \textbf{66.17} & \textbf{66.09} & \textbf{72.51} & \textbf{69.54} & \textbf{72.38}
\textcolor{red}{(\textbf{+28.29})} 
& \textbf{74.80} 
\textcolor{red}{(\textbf{+36.2})} 
& \textbf{73.57}
\textcolor{red}{(\textbf{+32.41})} 
\\
\bottomrule
\end{tabular}
}
\end{table*}

Following the GRPO training paradigm, the group-wise advantage of each sampled response is computed by
\begin{equation}
A_i = \frac{R_i - \mathrm{mean}(\{R_j\})}{\mathrm{std}(\{R_j\})},
\end{equation}
where \(\{R_j\}\) represents the rewards of all sampled responses within the same group. Let
\begin{equation}
\rho_i = \frac{\pi_{\theta}(o_i \mid q)}{\pi_{\theta_{\mathrm{old}}}(o_i \mid q)}.
\end{equation}
The final GRPO objective is formulated as
\begin{align}
\mathcal{J}_{\mathrm{GRPO}}(\theta) 
&= \mathbb{E}_{q, \{o_i\}} \left[ \frac{1}{N}\sum_{i=1}^N \mathcal{L}_i - \beta \, \mathbb{D}_{\mathrm{KL}} \bigl(\pi_{\theta} \| \pi_{\mathrm{ref}}\bigr) \right], \\
\mathcal{L}_i 
&= \min \Bigl( \rho_i A_i,\, \mathrm{clip}(\rho_i, 1-\epsilon, 1+\epsilon) A_i \Bigr).
\end{align}
% where \(q\) denotes the input query, \(o_i\) denotes the \(i\)-th sampled response, and \(\pi_{\mathrm{ref}}\) is the reference policy. This optimization encourages the model to generate captions that better balance factual correctness and descriptive coverage.
Here, \(q\) denotes the input query, \(o_i\) denotes the \(i\)-th sampled response, and \(\pi_{\mathrm{ref}}\) is the reference policy. This optimization encourages the model to generate captions that achieve a better balance between factual correctness and descriptive coverage.

\section{Experiments}

\subsection{Implementation Details}
During the SFT stage, we train the base model Qwen3-VL-8B \cite{Qwen2025Qwen3VL} on 80K samples for 2 epochs, with a batch size of 128 and a learning rate of $2\times10^{-5}$. During the GRPO stage, we further train on 2K samples for 1 epoch, using 8 rollouts, a learning rate of $1\times10^{-5}$, and a prompt-wise batch size of 32. In the reward design, we set $\lambda_{\mathrm{acc}}=0.5$, $\lambda_{\mathrm{cmp}}=0.5$, and $\lambda_{\mathrm{cov}}=0.1$, while the gate threshold ($\tau$) for activating the coverage reward is set to 0.75. Videos are sampled at 2 FPS, with a maximum of 256 tokens per frame. All experiments are conducted on $32\times 80$ GB GPUs. More implementation details are provided in the appendix.

\subsection{Comparison with State of the Art}
To evaluate the effectiveness of CineCap, we compare it against a broad set of strong baselines, including both proprietary and open-source multimodal models. Specifically, the proprietary baselines include Gemini-2.5-Pro \cite{gemini25pro} and Gemini-3.1-Pro \cite{gemini31pro}, while the open-source baselines include Qwen3-VL-30B \cite{Qwen2025Qwen3VL}, Qwen2.5-VL-72B \cite{bai2025qwen2}, Qwen3-VL-8B \cite{Qwen2025Qwen3VL}, Tarsier-7B \cite{yuan2024tarsier}, InternVL3-8B \cite{Zhu2025InternVL3}, LLaVA-OneVision-7B \cite{li2024llava}, and LLaVA-NeXT-Video-7B \cite{zhang2024llavanextvideo}. These baselines cover representative recent models with strong visual understanding and generation capabilities, enabling a comprehensive comparison on cinematographic captioning.

For evaluation, we report both \textbf{aspect-level} and \textbf{overall-level} results under the two criteria defined in Sec.\ref{eval}: \textbf{comprehensiveness} (Cmp) and \textbf{accuracy} (Acc). The aspect-level evaluation covers six cinematographic dimensions, including Camera Movement (CM), Shot Size (SS), Depth of Field (DF), Camera Angle (CA), Composition (CO), and Subject Orientation (SO), while the overall level further reports holistic Cmp, Acc, and F1.

Tab.~\ref{tab:main_result} shows that CineCap consistently outperforms all proprietary and open-source baselines across all metrics. In particular, CineCap achieves 72.38 overall Cmp, 74.80 overall Acc, and 73.57 F1, significantly surpassing the strongest baseline Gemini-3.1-Pro (46.31, 60.48, 52.45). This demonstrates that our method improves not only factual correctness but also descriptive coverage.

The gains are consistent across all six cinematographic dimensions. The improvements are especially large on Camera Movement, Shot Size, Depth of Field, and Composition, where multi-dimensional dense description requires both fine-grained visual understanding and explicit structured reasoning. For example, on Camera Movement, CineCap improves Cmp/Acc from 40.23/43.09 to 54.38/60.50, and on Depth of Field from 48.11/47.37 to 79.68/82.91. Meanwhile, the strong performance on Camera Angle (90.06/88.35) and Subject Orientation (72.51/69.54) further indicates that our framework can jointly support both dynamic and static cinematographic description within a unified caption. These results verify the effectiveness of our spatio-temporal structured reasoning and reward design for multi-dimensional cinematographic captioning.

\begin{table}[t]
\centering
\caption{Ablation study on the effectiveness of different fine-tuning strategies. We progressively add components to the Base model. \textbf{Acc}: Overall Accuracy, \textbf{Cmp}: Overall Comprehensiveness. The best results are highlighted in \textbf{bold}.}
\label{tab:ablation}
\begin{tabular}{l ccc c}
\toprule
Model & Cmp & Acc & F1 & $\Delta$ F1 \\
\midrule
Base (Qwen3)\cite{Qwen2025Qwen3VL} & 44.09 & 38.60 & 41.16 & - \\
\quad + Direct Caption SFT & 68.00 & 70.58 & 69.27 & \textcolor{green!60!black}{+28.11} \\
\quad + CineCap SFT & 69.36 & 71.08 & 70.21 & \textcolor{green!60!black}{+29.05} \\
\quad + CineCap SFT w/ GRPO & \textbf{72.38} & \textbf{74.80} & \textbf{73.57} & \textcolor{green!60!black}{\textbf{+32.41}} \\
\bottomrule
\end{tabular}
\end{table}

\begin{table}[t]
\centering
\caption{Ablation study on the reward design. Starting from the Baseline, we first add the \textit{Cmp \& Acc Reward}. We then compare three parallel strategies applied on top of this strong foundation. \textbf{Acc}: Overall Accuracy, \textbf{Cmp}: Overall Comprehensiveness. The best result in each column is highlighted in \textbf{bold}.}
\label{tab:reward_ablation}
\begin{tabular}{l ccc c}
\toprule
Model & Cmp & Acc & F1 & $\Delta$ F1 \\
\midrule
CineCap SFT & 69.36 & 71.08 & 70.21 & - \\
\quad + Cmp \& Acc Reward & 71.40 & 74.52 & 72.93 & \textcolor{green!60!black}{+2.72} \\
\midrule
\multicolumn{5}{l}{\textit{Parallel variations based on (+ Cmp \& Acc Reward):}} \\
\quad w/ Len Penalty & \textbf{73.29} & 70.20 & 71.71 & \textcolor{green!60!black}{+1.50} \\
\quad w/ Cov (w/o gate) & 72.27 & 73.96 & 73.11 & \textcolor{green!60!black}{+2.90} \\
\quad w/ Cov (w/ gate) & 72.38 & \textbf{74.80} & \textbf{73.57} & \textcolor{green!60!black}{\textbf{+3.36}} \\
\bottomrule
\end{tabular}
\end{table}

% \begin{findingbox}
% % \textbf{Finding.}
% Caption length is not a reliable proxy for cinematographic completeness. The best results are achieved by regulating atomic aspect coverage, and further applying a gate to preserve accuracy.
% \end{findingbox}
\subsection{Ablation Analysis}
\subsubsection{Fine-tuning Strategy Ablation}
Tab.~\ref{tab:ablation} shows the ablation of progressive fine-tuning strategies. From the base model, direct caption SFT raises overall F1 from 41.16 to 69.27, confirming that curated cinematographic caption data provide strong supervision. However, this gain is not solely due to data. Replacing direct caption SFT with CineCap SFT, which uses the same training source but adds spatio-temporal anchor-based atomic CoT supervision, further improves F1 from 69.27 to 70.21. This indicates that structured supervision, not just more data, drives this improvement.
Applying CineCap GRPO on top of CineCap SFT raises overall F1 from 70.21 to 73.57 with simultaneous gains in accuracy and comprehensiveness. This verifies the reward design effectively complements supervised learning by balancing descriptive coverage and factual correctness. The consistent improvements in both metrics suggest the gain arises from better alignment of captions with the cinematographic structure rather than longer or more aggressive generation.
\begin{findingbox}
Structured spatio-temporal atomic supervision improves performance beyond curated caption data alone. The proposed reward further enhances accuracy and comprehensiveness by balancing coverage and correctness, promoting better caption-structure alignment.
\end{findingbox}
\subsubsection{Reward Design Ablation}
Table~\ref{tab:reward_ablation} presents an analysis of reward design commencing from CineCap SFT. The addition of the Cmp \& Acc Reward increases the overall F1 score from 70.21 to 72.93, demonstrating the advantage of reinforcement learning on these objectives. Introducing a naive length penalty improves comprehensiveness but decreases accuracy, which causes the F1 score to decline to 71.71, indicating that longer captions tend to be more complete but less accurate. Substituting the length penalty with the proposed coverage reward yields improved balance between metrics, confirming that atomic statement coverage serves as a more effective control signal than caption length. The inclusion of the gating mechanism further enhances the final results to 74.80 in accuracy, 72.38 in comprehensiveness, and 73.57 in F1—the highest among all variants—validating that the gated coverage reward effectively balances descriptive completeness and factual accuracy.
\begin{findingbox}
Caption length is an inadequate proxy for cinematographic completeness. Controlling atomic aspect coverage through a gating mechanism attains a superior balance between accuracy and coverage.
\end{findingbox}
% Tab.~\ref{tab:reward_ablation} analyzes the reward design starting from CineCap SFT. Adding the Cmp \& Acc Reward raises overall F1 from 70.21 to 72.93, demonstrating the benefit of reinforcement learning on these objectives. Adding a naive length penalty increases accuracy but lowers comprehensiveness, resulting in a reduced F1 of 71.71 which implies shorter captions are safer but less complete. Replacing length penalty with the proposed coverage reward leads to improved trade-offs, confirming atomic statement coverage is a better control signal than caption length. Incorporating the gating mechanism further improves final results to 72.38 Acc, 74.80 Cmp, and 73.57 F1, the best among all variants, verifying gated coverage reward effectively balances descriptive completeness and factual correctness.
% \begin{findingbox}
% Caption length is a poor proxy for cinematographic completeness. Regulating atomic aspect coverage with a gating mechanism achieves better balance between accuracy and coverage.
% \end{findingbox}

\begin{figure}[t]
    \centering
    \includegraphics[width=\columnwidth]{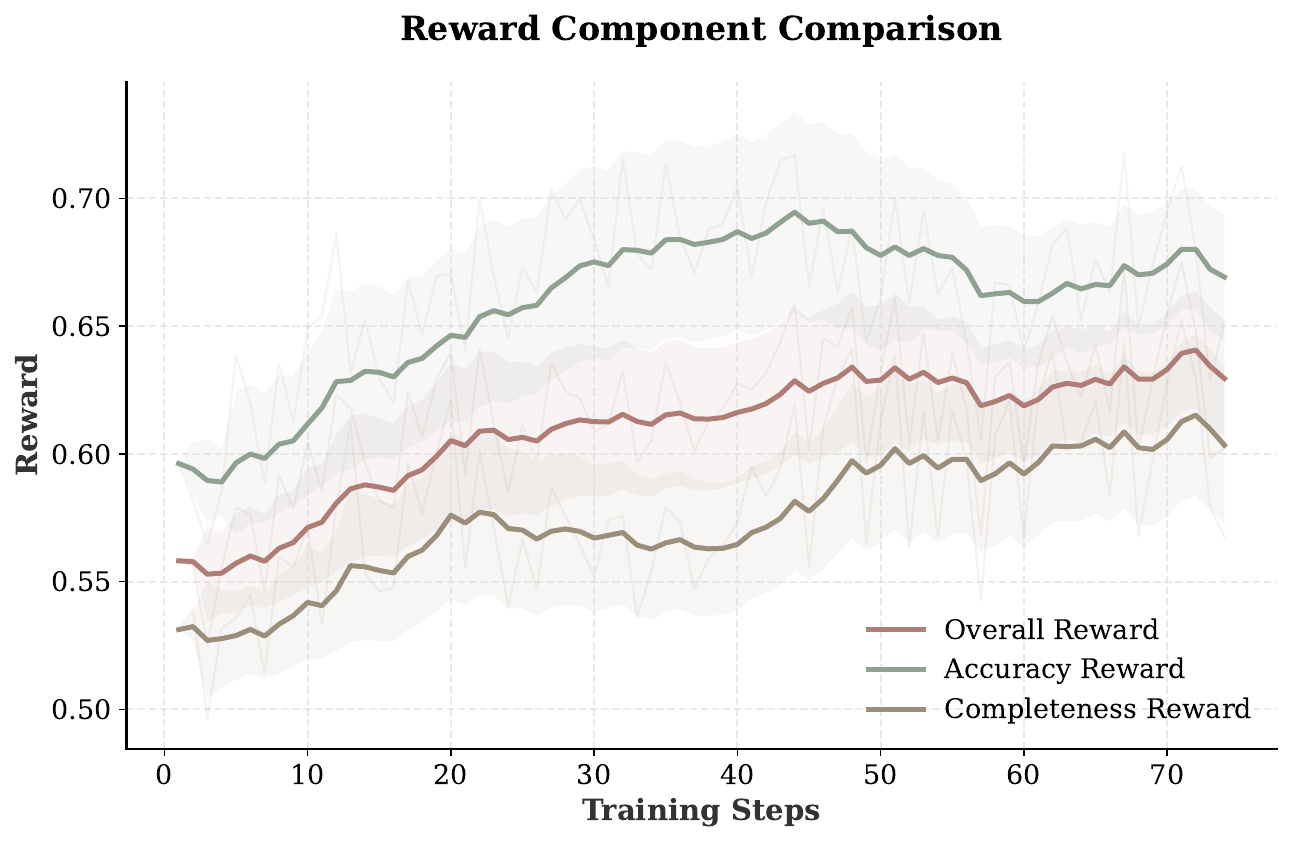}
    \caption{Training dynamics of reward components during reinforcement learning. All reward curves increase steadily, indicating stable optimization.} 
    \label{fig:reward_components}
\end{figure}

Fig.~\ref{fig:reward_components} shows the training dynamics of different reward components. All curves increase steadily, indicating stable GRPO optimization. The accuracy reward grows faster and remains higher than the completeness reward. Overall, the trend supports the effectiveness of our reward design.

\section{Conclusion}

This work proposes \ours, a unified framework for cinematographic captioning, and formulates the task as a multi-dimensional dense description problem across six essential cinematographic dimensions. The proposed method integrates spatio-temporal anchor-based structured reasoning, atomic chain-of-thought supervision, and GRPO-based reward optimization to enhance both descriptive completeness and factual accuracy. To enable systematic evaluation, CineCap Bench is introduced, offering fine-grained assessments at both the aspect and overall levels. We conduct extensive experiments showing that \ours consistently outperforms both open-source and proprietary baselines, achieving up to 32.41\% improvement in F1 evaluation. It is expected that this work will promote further research on cinematographic understanding and provide valuable support for downstream tasks involving controllable cinematic video generation.
\bibliographystyle{ACM-Reference-Format}
\bibliography{main}

% \appendix









\end{document}